\documentclass{article}

\usepackage{arxiv}

\newcommand{\fulltitle}{Uncertainty in Extreme Multi-label Classification}

\usepackage{algorithm}
\usepackage{algorithmic}

\usepackage{amsmath,amsfonts,amssymb,amsthm}

\usepackage{bbm}
\usepackage{bm}
\usepackage{enumitem}
\usepackage{multirow}
\usepackage{thmtools}
\usepackage{thm-restate}
\usepackage{xspace}

\newcommand{\mysection}[1]{{\noindent \bf #1.}\xspace}

\newcommand{\btheta}{\bm{\theta}\xspace}
\newcommand{\cD}{\mathcal{D}\xspace}
\newcommand{\cO}{\mathcal{O}\xspace}
\newcommand{\cH}{\mathcal{H}\xspace}
\newcommand{\cI}{\mathcal{I}\xspace}
\newcommand{\cL}{\mathcal{L}\xspace}
\newcommand{\cM}{\mathcal{M}\xspace}
\newcommand{\bx}{\bm{x}\xspace}
\newcommand{\bbE}{\mathbb{E}\xspace}
\newcommand{\bbV}{\mathbb{V}\xspace}

\newcommand{\eurlex}{{\sf Eurlex-4K}\xspace}
\newcommand{\wikis}{{\sf Wiki10-31K}\xspace}
\newcommand{\amzcat}{{\sf AmazonCat-13K}\xspace}
\newcommand{\amzsmall}{{\sf Amazon-670K}\xspace}
\newcommand{\amzlarge}{{\sf Amazon-3M}\xspace}
\newcommand{\wikil}{{\sf Wiki-500K}\xspace}

\newcommand{\tfidf}{tfidf\xspace}

\def\xrlinear{XR-LINEAR\xspace}
\def\xrtransformer{XR-TRANSFORMER\xspace}

\newcommand\numberthis{\addtocounter{equation}{1}\tag{\theequation}}

\newtheorem{theorem}{Theorem}[section]

\newtheorem{lemma}[theorem]{Lemma}

\usepackage[utf8]{inputenc} 
\usepackage[T1]{fontenc}    
\usepackage{hyperref}       
\usepackage{url}            
\usepackage{booktabs}       
\usepackage{amsfonts}       
\usepackage{nicefrac}       
\usepackage{microtype}      
\usepackage{cleveref}       
\usepackage{lipsum}         
\usepackage{graphicx}
\usepackage[sort,square,numbers]{natbib}
\usepackage{doi}

\title{\fulltitle}

\author{
    {Jyun-Yu Jiang}\\
    Amazon Search\\
    Palo Alto, CA 94301\\
    \texttt{jyunyu.jiang@gmail.com}
    \And
    {Wei-Cheng Chang}\\
    Amazon Search\\
    Palo Alto, CA 94301\\
    \texttt{weicheng.cmu@gmail.com}
    \And
    {Jiong Zhang}\\
    Amazon Search\\
    Palo Alto, CA 94301\\
    \texttt{zhangjiong724@gmail.com}
    \And
    {Cho-Jui Hsieh}\\
    University of California, Los Angeles\\
    Los Angeles, CA 95005\\
    \texttt{chohsieh@cs.ucla.edu}
    \And
    {Hsiang-Fu Yu}\\
    Amazon Search\\
    Palo Alto, CA 94301\\
    \texttt{rofu.yu@gmail.com}
}

\renewcommand{\shorttitle}{\fulltitle}

\hypersetup{
pdftitle={\fulltitle},
pdfsubject={cs.ML},
pdfauthor={Jyun-Yu Jiang, Wei-Cheng Chang, Jiong Zhong, Cho-Jui Hsieh, Hsiang-Fu Yu},
pdfkeywords={Extreme multi-label classification, Uncertainty quantification}
}

\begin{document}
\maketitle

\begin{abstract}
	Uncertainty quantification is one of the most crucial tasks to obtain trustworthy and reliable machine learning models for decision making.
However, most research in this domain has only focused on problems with small label spaces and ignored eXtreme Multi-label Classification (XMC), which is an essential task in the era of big data for web-scale machine learning applications.
Moreover, enormous label spaces could also lead to noisy retrieval results and intractable computational challenges for uncertainty quantification.
In this paper, we aim to investigate general uncertainty quantification approaches for tree-based XMC models with a probabilistic ensemble-based framework.
In particular, we analyze label-level and instance-level uncertainty in XMC, and propose a general approximation framework based on beam search to efficiently estimate the uncertainty with a theoretical guarantee under long-tail XMC predictions.
Empirical studies on six large-scale real-world datasets show that our framework not only outperforms single models in predictive performance, but also can serve as strong uncertainty-based baselines for label misclassification and out-of-distribution detection, with significant speedup.
Besides, our framework can further yield better state-of-the-art results based on deep XMC models with uncertainty quantification.

\end{abstract}


\section{Introduction}
\label{section:intro}

Extreme multi-label classification (XMC), or extreme multi-label learning, aims to find the relevant labels for a data input from an enormous label space.
With increasingly growing information in the era of big data, XMC has become more and more important, and has been widely applied to various real-world applications, such as advertising~\cite{prabhu2018parabel}, product search~\cite{chang2021extreme}, and document retrieval~\cite{bhatia2015sparse}.
However, for domains with potential high risks from mistakes like public health and medicine, it is crucial to model the predictive uncertainty for their downstream XMC applications like food classification~\cite{zheng2014spectroscopy} and medical diagnosis~\cite{almagro2020icd}.
In particular, an input sometimes could have only few or even no matches in the label space, so the outputs could be noisy without uncertainty quantification.
It is also insufficient to only model uncertainty for the entire input since XMC models could have different confidence for each label among the whole enormous space.

To estimate predictive uncertainty, Bayesian and probabilistic models~\cite{ghahramani2015probabilistic} are inherently applicable because variance can intrinsically be viewed as an uncertainty measurement.
However, although Bayesian approaches are mathematically grounded to model uncertainty, their computational costs are usually exorbitant for large-scale data.
To address this issue, the most popular solution is to approximate Bayesian inference by sampling models as an ensemble~\cite{Gal2016Uncertainty}.
Accordingly, ensemble-based Bayesian approximation has been applied to analyze the uncertainty in neural networks~\cite{gal2016dropout,depeweg2018decomposition}, gradient boosting based on decision trees~\cite{malinin2020uncertainty}, autoregressive structured prediction~\cite{malinin2019uncertainty}, and random forests~\cite{shaker2020aleatoric}.
Unfortunately, none of the existing studies has studied uncertainty quantification for XMC models.

Uncertainty quantification for XMC models is challenging.
First, different from conventional classification models, XMC models usually focus on deriving a small subset of relevant labels from the enormous label space.
In other words, for XMC models, we not only should consider how confident a model is for the entire input, but also need to model the uncertainty for each individual label.
However, most of the existing uncertainty quantification studies only concentrate on instance-level uncertainty~\cite{abdar2021review}.
The enormous number of labels could also result in computational difficulty.
Second, XMC models usually take extremely sparse input features~\cite{you2019attentionxml,chang2021extreme} with distillation~\cite{yu2020pecos}.  
Moreover, many XMC models~\cite{prabhu2018parabel,yu2020pecos} conduct convex optimization for better computational efficiency, which are insensitive to initialization~\cite{ruo2009smoothing}.
As a result, existing uncertainty quantification approaches that manipulate model weights, such as MC Dropout~\cite{gal2016dropout} and Deep Ensemble~\cite{lakshminarayanan2017simple}, could be ineffective.

To address these issues, in this work, we investigate ensemble-based uncertainty quantification for XMC models.
We first present the concept of modeling \emph{label-level} and \emph{instance-level} uncertainty.
To tackle the efficiency issue due to enormous labels, we propose a general framework to approximate uncertainty measurements by beam search with a theoretical guarantee under long-tail probability distributions.
Our contributions can be summarized as:
\begin{itemize}[leftmargin=*]
    \item We are the pioneer of uncertainty quantification for XMC models. Especially, we broaden the scope by simultaneously modeling \emph{label-level} and \emph{instance-level} uncertainty, which is an essence of XMC tasks.

    \item We propose an efficient and general framework to approximate uncertainty in XMC.
    With the observation of long-tail distributions in predictions, beam search on tree-based XMC models can further mitigate the computational efficiency with mathematical guarantees on estimated uncertainty.
    \item We conduct experiments on six benchmark XMC datasets. Our approaches can not only obtain satisfactory predictive performance, but also appropriately estimate both \emph{label-level} and \emph{instance-level} uncertainty. Besides, using \xrtransformer as deep base models, our framework can obtain better state-of-the-art results.
\end{itemize}

\section{Preliminaries}
\label{section:prelim}


\subsection{Ensemble-based Uncertainty Quantification}
\label{section:prelim:ensemble}

In this work, we focus on ensemble-based uncertainty quantification using Bayesian ensembles~\cite{Gal2016Uncertainty}, which learns the ensemble of multiple individual models.
The model parameters $\btheta$ are considered as random variables from a posterior distribution $p(\btheta\mid \cD)$, which can be computed by the Bayes' rule as:
\begin{equation*}
p(\btheta\mid \cD) = \frac{p(\cD\mid \btheta) p(\btheta)}{p(\cD)}
\end{equation*}
where $\cD = \lbrace \bx_i, y_i \rbrace$ is the training dataset, and $p(\cD)$ is the prior data distribution.
More precisely, each set of model parameters $\btheta$ is a sample from the posterior $p(\btheta\mid \cD)$ to demonstrate a hypothesis~\cite{bernardo2009bayesian} learned from the observations presented by the training data $\cD$.

Since the exact Bayesian inference could be intractable, a conventional approach is to consider an approximated distribution $q(\btheta)$ and mimic the true posterior $p(\btheta\mid \cD)$~\cite{bernardo2009bayesian}.
Specifically, exploiting ensemble models is one of the most popular approximation methods~\cite{chipman2007bayesian}.
Suppose we have an ensemble of $M$ probabilistic models $\lbrace \Pr(y\mid \bx; \btheta^{(m)})\rbrace_{m=1}^M$ sampled from the posterior $p(\btheta\mid \cD)$.
The \emph{predictive posterior} $\Pr(y\mid \bx, \cD)$ for inference approximation based on the ensemble can be estimated by computing the expectation over the ensemble models as:
$$\Pr(y\mid \bx, \cD) = \int_{\btheta} p(y\mid \bx; \btheta) p(\btheta \mid \cD) d\btheta \approx \bbE_{q(\btheta)}[\Pr(y\mid \bx; \btheta)] \approx \frac{1}{M} \sum_{m=1}^M \Pr(y\mid \bx; \btheta^{(m)}),$$
where $\btheta^{(m)} \sim q(\btheta) \approx p(\btheta\mid \cD)$ represents the model parameters of each individual model in the ensemble.

\mysection{Uncertainty Quantification via Entropy}
For a probabilistic model and its outputs, entropy is a native way to estimate the uncertainty~\cite{wehrl1978general}.
Given the predictive posterior $\Pr(y\mid \bx, \cD)$, the overall uncertainty, or so-called \emph{total uncertainty}~\cite{depeweg2018decomposition}, can be estimated as:
\begin{equation}\label{eq:totalunc}
    \cH[\Pr(y\mid \bx, \cD)] = \bbE_{p(y \mid \bx, \cD)}[-\ln \Pr(y \mid \bx, \cD)] \approx \cH[\frac{1}{M} \sum_{m=1}^M \Pr(y\mid \bx; \btheta^{(m)})].
\end{equation}
However, \emph{total uncertainty} incorporates both epistemic (\emph{knowledge}) uncertainty and aleatoric (\emph{data}) uncertainty~\cite{depeweg2018decomposition, kirsch2019batchbald}.
Previous studies also demonstrate that \emph{knowledge uncertainty} could be more beneficial for downstream applications, such as active learning~\cite{kirsch2019batchbald} and out-of-distribution detection~\cite{charpentier2020posterior}.
To compute the \emph{knowledge uncertainty}, we can decompose the \emph{total uncertainty} by deriving the mutual information between the model parameters $\btheta$ and the prediction $y$~\cite{depeweg2018decomposition} as:
\begin{align*} \numberthis \label{eq:tuku}
    & \underbrace{\cI[y, \btheta\mid \bx, \cD]}_{\emph{Knowledge Uncertainty}} = \underbrace{\cH[\Pr(y\mid \bx, \cD)]}_{\emph{Total Uncertainty}} - 
    \underbrace{\bbE_{p(\btheta\mid \cD)}[\cH[\Pr(y\mid \bx; \btheta)]]}_{\emph{Expected Data Uncertainty}}\\
   &  \approx \cH[\frac{1}{M} \sum_{m=1}^M \Pr(y| \bx; \btheta^{(m)})] - \frac{1}{M} \sum_{m=1}^M \cH[\Pr(y| \bx; \btheta^{(m)})].
\end{align*}

\mysection{Uncertainty Quantification via Variation}
If we treat the inference process as a regression problem to derive continuous probabilities, the variation over predicted probabilities of individual models in the ensemble can be considered as another direction to estimate the uncertainty~\cite{shelmanov2021certain}. Specifically, the uncertainty can be estimated by computing the \emph{probability variance} as:
\begin{equation} \label{eq:var}
\bbV_{p(y\mid \bx, \cD)}[\Pr(y\mid \bx; \btheta)] \approx \bbV_{q(\btheta)}[\Pr(y\mid \bx; \btheta)]\approx \frac{1}{M} \sum_{m=1}^M [ \Pr(y\mid \bx; \btheta^{(m)}) - \mu  ]^2, 
\end{equation}
where $\mu = \frac{1}{M} \sum_{m=1}^M \Pr(y\mid \bx; \btheta^{(M)})$ is the mean predicted probability of all individual models in the ensemble.


In this work, we examine three quantification approaches in our experiments, including \emph{total uncertainty} (TU), \emph{knowledge uncertainty} (KU), and \emph{probability variance} (PV).

\subsection{eXtreme Multi-label Classification (XMC)}
\label{section:prelim:xmc}

Given a training dataet $\cD$ of $N$ instances, for an arbitrary testing instance $\bm{x}$, an XMC model aims to estimate a probability $\Pr(y_\ell \mid \bm{x}, \cD)$ for each label $l\in\mathcal{L}$ in an extreme label space $\mathcal{L}$ with $L$ labels.

\mysection{Training with Negative Example Selection}
With the extreme label space, it is time-consuming for models to be trained with all negative examples.
As a result, most of the state-of-the-art XMC models select appropriate negative examples during optimization.
For example, Parabel~\cite{prabhu2018parabel} and PECOS~\cite{yu2020pecos} consider teacher forcing negatives~\cite{lamb2016professor} and matcher-aware negatives for better performance.
For simplicity, we denote the adjusted training dataset with negative example selection as $\cD^\prime$ so that the targets of those XMC models become $\Pr(y_\ell \mid \bx, \cD^\prime) \approx \Pr(y_\ell \mid \bx, \cD)$.

\begin{figure*}[!t]
    \includegraphics[width=\linewidth]{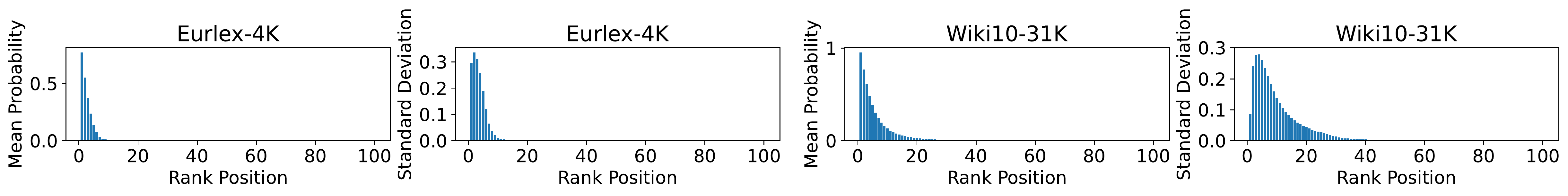}
    \caption{The mean and standard deviation of predicted probabilities over different rank positions on the testing data of \textsf{Eurlex-4K} and \textsf{Wiki10-31K} datasets.}
    \label{fig:longtail}
\end{figure*}

\mysection{Hierarchical Label Tree and Semantic Indexing}
Due to the enormous labels, it can be inappropriate to have linear inference time complexity to the number of labels.
To achieve acceptable efficiency, one of the most prominent methods is to partition the enormous label space into a hierarchical label tree~\cite{yu2020pecos, prabhu2018parabel, khandagale2020bonsai}.
Specifically, top-down $B$-ary clustering can recursively construct a depth-$d$ hierarchical label tree with label representations like instance feature aggregation and additional label features.
The cluster numbers $K_t$ are $B^{t}$ and $L$ for the first $d-1$ layers and the last $d$-th layer.
The clustering assignment at each layer $t$ can then be represented by an indexing matrix $C^{(t)}\in \lbrace 0, 1\rbrace ^{K_t \times K_{t-1}}$ as :
\begin{equation*}
C^{t}_{\ell k} = \left\{ \begin{array}{ll}1, & \text{if } k = c^t_\ell\\ 0, & \text{otherwise} \end{array} \right.,
\end{equation*}
where $c^{(t)}_{\ell k}$ is the corresponding cluster at the layer $t-1$ for a cluster (or a label when $t=d$) at the layer $t$ in the hierarchical label tree.

With the hierarchical label tree and the indexing matrix $C^{(t)}$, tree-based XMC models can conduct semantic indexing with beam search~\cite{kumar2013beam} for efficient approximated inference in a sub-linear time.
More precisely, for every clustering layer $t-1$, tree-based models estimate $\Pr(y^{t}_\ell \mid \bx, \cD^{\prime(t)})$ and encode top-$b$ clusters into $Y^{(t-1)}\in\{0,1\}^{K_{t-1}}$, where $\cD^{\prime(t)}$ is the induced training dataset for clusters at the layer $t$; $\sum Y^{(t-1)}[i] = b$.
Models can then focus on limited clusters or labels indicated by $Y^{(t-1)}C^{t} \in \{0, 1\}^{K_t}$ for inference at the clustering layer $t$.
Finally, $\Pr(y_\ell\mid \bx, \cD^\prime)$ can then be computed as:
\begin{equation*}
\hspace{-6pt} \Pr(y_\ell\mid \bx, \cD^\prime) = \Pr(y^{d}_\ell \mid \bx, \cD^{\prime(d)}) \times \Pr(c^d_\ell \mid \bx, \cD^{\prime(d-1)}), 
\end{equation*}
where $\Pr(c^d_\ell \mid \bx, \cD^{\prime(t-1)})$ can be further recursively derived through clustering layers.
In this paper, we denote a tree-based XMC model learn a parameter set $\btheta$ from the dataset to recursively compute the predictions as:
\begin{equation}\label{eq:treerec}
    \Pr(y_\ell\mid \bx; \btheta) = \Pr(y^{d}_\ell \mid \bx, c^d_\ell; \btheta) \times \Pr(c^d_\ell \mid \bx; \btheta). 
\end{equation} 

\mysection{Challenges in XMC Uncertainty Quantification}
Although the uncertainty measures introduced in Section~\ref{section:prelim:ensemble} are tractable in most scenarios, a linear computational time over all enormous labels could still be unacceptable.
Besides, uncertainty for XMC models can be examined in two levels, including (1) \emph{label-level} for the predicted probability $\Pr(y_\ell \mid \bx, \cD)$ of each single label $l$, and (2) \emph{instance-level} for the entire instance $\bx$.
However, none of the existing studies has addressed any of these two uncertainty types for XMC models.

\section{XMC Uncertainty Quantification}
\label{section:method}

In this section, we propose an efficient framework to approximate both label-level and instance-level uncertainty.
We also suggest two approaches to generate model ensembles.


\subsection{Uncertainty in XMC}
\label{section:method:uncertainty}

Here we propose to model two different levels of uncertainty, including \emph{label-level} and \emph{instance-level} uncertainty.
The former represents the confidence of the decision on each individual label, and the latter estimates how certain the model is for a given data instance.


\mysection{Label-level Uncertainty}
Label-level uncertainty quantification can be reduced to $L$ subtasks of estimating uncertainty of binary classification for each label $\ell\in \cL$.
Specifically, given a model ensemble, we collect $M$ predicted probabilities $\lbrace \Pr(y_\ell \mid \bx; \btheta^{(m)}\rbrace_{m=1}^M$ for each label $\ell$, thereby estimating the uncertainty based on different approaches introduced in Equations~\eqref{eq:totalunc}, \eqref{eq:tuku}, and \eqref{eq:var} in Section~\ref{section:prelim:ensemble}.

\mysection{Instance-level Uncertainty}
To estimate instance-level uncertainty, it is required to simultaneously consider the predictions over all labels $\ell\in \cL$ for an instance $\bx$.
As a pioneer of this direction in XMC, we borrow the idea of \emph{JointEnergy}~\cite{wang2021can, lecun2006tutorial} in the domain of conventional multi-label classification by adding up uncertainty metrics over all labels.
Specifically, from the joint likelihood perspective, instance-level \emph{Total Uncertainty}, which estimates the entropy for each label, can be decomposed into an explainable form under the assumption of conditional independence for $p(y_\ell\mid \bx)$ as:
\begin{align*}
     \sum_{\ell\in \cL} \cH[p(y_\ell)] 
    & = \sum_{\ell\in \cL} - \ln p(y_\ell\mid \bx) \\
    & = -\sum_{\ell\in \cL} \ln p(\bx\mid y_\ell) - \sum_{\ell\in \cL} Z_\ell\\
    & = -\ln \prod_{\ell\in \cL} \frac{p(y_\ell\mid \bx)p(\bx)}{p(y_\ell)} - \sum_{\ell\in \cL} Z_\ell\\
    & = -\ln p(y_1, \dots, y_L\mid \bx)  - L \ln p(\bx) + \ln \prod_{\ell\in \cL} p(y_\ell) - \sum_{\ell\in \cL} Z_\ell \\
    & =  -\ln \frac{p(\bx \mid y_1, \dots, y_L) \prod_{\ell\in \cL} p(y_\ell)}{p(\bx)} - L \ln p(\bx)  + \ln \prod_{\ell\in \cL} p(y_\ell) - \sum_{\ell\in \cL} Z_\ell \\
    & = \underbrace{-\ln p(\bx\mid y_1, \dots y_L)}_{\text{Knowledge Uncertainty}} \underbrace{-(L-1)\ln p(\bx)}_{\text{Data Uncertainty}} - \sum_{\ell\in \cL} Z_\ell, 
\end{align*}
where $Z_\ell = p(\bx\mid y_\ell) / p(y_\ell\mid \bx)$ is the normalized density for each label $\ell\in \cL$. 

\subsection{Uncertainty~Approximation~via~Beam~Search}

Even though we now have tractable approaches to estimate both label-level uncertainty and instance-level uncertainty, their linear computational time could still be too slow to address XMC tasks that usually have millions of labels.
In this work, we propose to leverage the power of long-tail probabilistic distributions and variances in XMC problems.

Figure~\ref{fig:longtail} illustrates the statistics of predicted probabilities over different rank positions on the testing data of the \textsf{Eurlex-4K} and \textsf{Wiki10-31K} datasets.
Both mean and standard deviation of predicted probabilities are long-tail over rank positions while most of the labels in the space are predicted with near-zero probabilities and variance.
In other words, the label-level uncertainty of those labels and the corresponding components in the computations of instance-level uncertainty would be also close to zero. 
Indeed, instances in XMC tasks tend to have very few labels, so the phenomenon is also intuitive.
Based on this observation, we propose to approximate uncertainty based on beam search and retrieving top-ranked but limited labels.

As mentioned in Section~\ref{section:prelim:xmc}, beam search~\cite{kumar2013beam} is a convenient tool to efficiently derive the top-ranked labels.
We also notice that the long-tail phenomenon on intermediate probabilities is also consistent through all layers in the search.
Hence, for uncertainty approximation, we propose to only consider the predicted probabilities of top-ranked labels returned by beam search and assign zero probabilities for the remaining labels.
The computational time of uncertainty can then be improved from linear to sub-linear in the number of labels.
In Algorithm~\ref{alg:approxbeam}, we present the pseudo code of the detail process for uncertainty approximation via beam search for a testing instance $\bx$.

\begin{algorithm}[tb]
\caption{Uncertainty Approximation via Beam Search}
\label{alg:approxbeam}
\begin{algorithmic}
    \STATE {\bfseries Input:} 
        a model ensemble $\lbrace \Pr(y\mid \bx; \btheta^{(m)})\rbrace_{m=1}^M$;
        indexing matrices $\lbrace C^{(t)}\rbrace_{1\leq t\leq d}$;
        the testing instance $\bx$.
    \STATE {\bfseries Output:} Approximated Uncertainty Metrics $\hat{U}$.
    \STATE {\bfseries Hyper-parameters:} the beam width $b$; the number of top-ranked labels $k$.
    \STATE labelProb = $\lbrace\delta\rbrace^{L\times M}$, where $\delta\approx 0$ is a small positive constant.
    \FOR{$m = 1$ {\bfseries to} $M$}
        \STATE Let $A = \mathbf{1}^1$ denote label/cluster candidates over layers.
        \FOR{$t = 1$ {\bfseries to} $d$}
            \STATE $A = \text{Binarize}({C^{(t)}}^\intercal \times A)$
            \STATE Calculate $\Pr(y^t_\ell|\bx; \btheta^{(m)})$ for each candidate $\ell$ in $A$.
            \STATE $k_t = b$ {\bfseries if} $t < d$ {\bfseries else} $k$
            \STATE $A = \mathbf{0}^{K_t}$
            \STATE $A[\ell$ in top-$k_t$ candidates$] = 1$
        \ENDFOR
        \FOR{top-ranked $\ell$ {\bfseries in} $A$}
            \STATE labelProb$[\ell, m]$ = $\Pr(y_\ell\mid \bx; \btheta^{(m)})$
        \ENDFOR
    \ENDFOR
    \STATE Compute $\hat{U}$ as approximated uncertainty with labelProb.
    \STATE {\bfseries return} $\hat{U}$
\end{algorithmic}
\end{algorithm}

\mysection{Theoretical Analysis}
In addition to the efficiency, the long-tail property also provides the theoretical guarantee on the accuracy of beam search results.
\begin{restatable}{theorem}{beamsearch}{} \label{theorem:beamsearch}
    Suppose $\Pr(y^t_\ell\mid \bx; \btheta)$ given by a tree-based model for each layer $t$ is under a long tail distribution as shown in Figure~\ref{fig:longtail}.
    With a large enough beam size $k^t$, the average regret of beam search in each layer $t$ would satisfy the following inequality:
    \begin{equation*}
        \frac{1}{k^t} \sum_{i=1}^{k^t} (\Pr(y^t_{o_i}\mid \bx; \btheta) -  \Pr(y^t_{\ell_i}\mid \bx; \btheta)) \leq \delta^\prime,
    \end{equation*}
    where $\delta^\prime \approx 0$ is a small positive constant; $o_i$ and $\ell_i$ are the oracle and predicted top-$i$ label.
\end{restatable}
The proof of Theorem~\ref{theorem:beamsearch} is presented in Appendix~\ref{appendix:theorem:beamsearch}.
The effectiveness of approximated \emph{total uncertainty} can be also shown in  Theorem~\ref{theorem:approx}
\begin{restatable}{theorem}{thmapprox}{} \label{theorem:approx}
    Suppose $\hat{U}(\bx, \ell)$ is the approximated total uncertainty given by Algorithm~\ref{alg:approxbeam} to estimate label-level total uncertainty for an instance~$\bx$ and each label~$\ell$; $\Pr(y_\ell\mid \bx, \btheta^{(m)})$ is under a long-tail distribution as shown in Figure~\ref{fig:longtail}.
    With a large enough hyper-parameter~$k$ in beam search, for each individual model, the approximated uncertainty $\hat{U}(\bx)$ and $\hat{U}(\bx, \ell)$ would satisfy the following inequality:
    \begin{equation*}
        |U(\bx, \ell) - \hat{U}(\bx, \ell)| \leq \delta^\prime,
    \end{equation*}
    where $\delta^\prime\approx 0$ is a small positive constant.
\end{restatable}
Note that we show the proof of Theorem~\ref{theorem:approx} in Appendix~\ref{appendix:theorem:approx}.
Other uncertainty metrics like \emph{knowledge uncertainty} and \emph{probability variance} can also reach similar properties.

\mysection{Time Complexity Analysis}
Suppose $d$ is the depth of tree-based models in the ensemble;
matrix multiplication is implemented with sparse matrices;
with $B$-ary clustering, the cluster number $K_t$ is $\min(B^t, L)$, where $B$ is a small constant; $T_{\btheta}$ is the time to compute $\Pr(y\mid \bx; \btheta)$. For simplicity, we let $k=b$; $T_{\btheta^{(m)}} = T_{\btheta}$ for all individual models.
The time complexity of Algorithm~\ref{alg:approxbeam} is $\cO\left(M d b T_{\btheta} \max\left(B, \frac{L}{B^{d-1}}\right)\right)$.
If $B$ and $d$ are decided such that $d = \cO(\log_B L)$ (i.e., $\frac{L}{B^{d-1}}$ is a small constant), the overall time complexity would be $\cO(M  v b T_{\btheta} \log L)$.
Compared to the complexity of na\"ive approach to compute the predictions over all labels for each model $\cO(M\times L\times T_{\btheta})$, our algorithm significantly reduces the computational time from linear to sub-linear to the extremely large size of the label spaces $L$.

\subsection{Generating XMC Model Enembles}
\label{section:method:generation}

Our framework is general for arbitrary methods of generating model ensembles.
In this work, as examples, we adopt two ensemble generation approaches from data and model perspectives without manipulating model weights, including bagging and boosting.

From the data aspect, bootstrap aggregating (\emph{bagging}) is one of the most popular ensemble meta-algorithms to enhance the stability of machine learning models~\cite{breiman1996bagging}.
In other words, individual models in the bagging ensemble can be representative of the diversity of model predictions so that the ensemble model can reduce variance~\cite{ganjisaffar2011bagging, domingos1997does}.
In addition to deriving ensemble models from the data perspective, the model perspective could be also important, so \emph{boosting} models~\cite{schapire2003boosting} can also appropriately generate model ensembles.
Specifically, we iteratively boost XMC models by leveraging earlier models to derive hard negatives as new training samples.
Moreover, \emph{Bagging} and \emph{Boosting} can be combined as \emph{Boosted Bagging} by deriving boosting models based on bootstrapped data and corresponding hard negatives.

\section{Experiments}
\label{section:exp}

\subsection{Experimental Settings}

\mysection{Datasets}
In this paper, we adopt six public benchmark extreme multi-label text classification datasets~\cite{you2019attentionxml}, including \textsf{Eurlex-4k}, \textsf{Wiki10-31K}, \textsf{Amazon-670K}, \textsf{AmazonCat-13K}, \textsf{Wiki-500K}, and \textsf{Amazon-3M}, as shown in Table~\ref{table:shortdatasets}.
We use the sparse \tfidf representations as features and training/testing data splits, which are consistent with existing studies in this field~\cite{you2019attentionxml,chang2020xmctransformer,zhang2021fast,jiang2021lightxml}.

\begin{table}[!h]
\centering
\caption{The statistics of six experimental datsets. Note that $n_{\text{train}}, n_{\text{test}}$ are the numbers of training and testing instnaces. 
        $L$ is the number of labels while $\bar{L}$ the average number of labels per instance.
        $\bar{n}$ the average number of instances per label.
}
\label{table:shortdatasets}
    \begin{tabular}{c|rrrrrrrr}
    \toprule
        Dataset         & $n_{\text{train}}$ & $n_{\text{test}}$   & $d$       & $L$       & $\bar{L}$ & $\bar{n}$ \\
        \midrule
        \eurlex         &       15,449       &           3,865     &   186,104 &     3,956 &     5.30  &     20.79 \\
        \wikis          &       14,146       &           6,616     &   101,938 &    30,938 &    18.64  &      8.52 \\
        \amzsmall       &      490,449       &         153,025     &   135,909 &   670,091 &     5.45  &      3.99 \\
        \amzcat         &    1,186,239       &         306,782     &   203,882 &    13,330 &     5.04  &    448.57 \\
        \wikil          &    1,779,881       &         769,421     & 2,381,304 &   501,070 &     4.75  &     16.86 \\
        \amzlarge       &    1,717,899       &         742,507     &   337,067 & 2,812,281 &    36.04  &     22.02 \\
        \bottomrule
        \end{tabular}
\end{table}

\mysection{Base XMC Models}
We first consider \xrlinear in PECOS~\cite{yu2020pecos} with sparse \tfidf representations as our base model since \xrlinear is one of the most popular tree-based XMC models with high flexibility and scalability to enormous output spaces and applied to various domains like extreme text classification~\cite{liu2021label} and product search~\cite{chang2021extreme}. 
We then adopt \xrtransformer~\cite{zhang2021fast}, one of the state-of-the-art text XMC models, to show the potential of our framework to be applied in deep XMC models.

\mysection{Evaluation Tasks and Metrics}
We consider three evaluation tasks: (1) predictive performance, (2) misclassification detection, and (3) out-of-distribution (OOD) deteciton.
Task~1 evaluates prediction accuracy with precision and recall metrics on top-ranked labels.
Task~2 assesses label-level uncertainty by detecting incorrectly ranked labels with uncertainty scores.
Task~3 appraises instance-level uncertainty by identifying testing instances out of training distributions with uncertainty scores.
Tasks~2 and 3 utilize area under the ROC curve (AUROC)~\cite{hendrycks2016baseline} as the evaluation metric.

\mysection{Baselines}
We mainly compare with single models.
For Tasks~2 and 3, we adopt \emph{Energy}~\cite{liu2020energy} and \emph{JointEnergy}~\cite{wang2021can}, the state-of-the-art OOD detection methods for conventional multi-class and multi-label classification, to derive baseline uncertainty scores.
Besides, since our approximation framework can be applied to arbitrary model ensembles.
We consider Monte-Carlo dropout (\emph{MC Dropout})~\cite{gal2016dropout} with a 5\% dropout rate as a comparative baseline to verify the effectiveness of our ensemble generation.

\mysection{Experimental Details}
Experiments described in Sections~\ref{section:exp:task1}, \ref{section:exp:task2}, \ref{section:exp:task3}, and \ref{section:exp:efficiency} without a need of GPUs are conducted on an AWS {\sf x1.32xlarge} instance with 128 CPU cores based on four Intel Xeon E7-8880 v3 \@ 2.30GHz processors and 1,952 GiB memory.
Experiments described in Section~\ref{section:exp:deep} with needs of GPUs are conducted on an AWS {\sf p3dn.24xlarge} instance with 96 CPU cores based on four Intel Xeon Platinum 8175M \@2.50GHz processors, 96 GiB memory, and 8 NVIDIA V100 Tensor Core GPUs with 32 GB of memory each.

For XR-Linear as our base XMC model, we establish hierarchical label trees based on Positive Instance Feature Aggregation (PIFA)~\cite{yu2020pecos} and 8-means (i.e., $B_t = 8$).
After training, model weights smaller than $1e-3$ are ignored for reducing disk space consumption, following the settings of \xrlinear~\cite{yu2020pecos}.
For inference, the beam size $b$ and the number of top-ranked labels $k$ in beam search are set as 50 and 100.
For the \emph{Bagging} ensemble approach, we bootstrap datasets with the identical size to the training dataset for training individual models.
For the \emph{Boosting} ensemble approach, we set the hyper-parameter $\alpha=0.5$.
For all ensemble-based approaches in the experiments (i.e., all methods except the single model), the number of model ensembles $M$ is 10 for ensemble generation.
All deep learning models using transformers use the {\sf bert-base-uncased} model as the pretrained model~\cite{devlin2018bert}.


\begin{table*}[!t]
\centering 
\resizebox{\textwidth}{!}{
\begin{tabular}{c|cccccc|cccccc|cccccc}
    \cline{2-19}
 & P@1 & P@3 & P@5 & R@1 & R@3 & R@5 & P@1 & P@3 & P@5 & R@1 & R@3 & R@5 & P@1 & P@3 & P@5 & R@1 & R@3 & R@5\\ \hline 
    Dataset  & \multicolumn{6}{c|}{\textsf{Eurlex-4K}} & \multicolumn{6}{c}{\textsf{Wiki10-31K}}  & \multicolumn{6}{c}{\textsf{Amazon-670K}} \\ \hline
  Single Model & 81.76 & 69.02 & 57.61 & 16.52 & 41.00 & 55.99 & 84.05 & 73.06 & 64.09 & 4.96 & 12.73 & 18.33 & 44.25 & 39.34 & 35.70 & 9.21 & 22.82 & 33.46\\\hline
  MC Dropout   & 81.91 & 68.95 & 57.67 & 16.55 & 40.95 & 56.07 & 84.04 & 73.06 & 64.07 & 4.96 & 12.74 & 18.33 & 44.29 & 39.37 & 35.70 & 9.21 & 22.84 & 33.47\\ 
  Boosting     & \bf 82.74 & 69.55 & \bf 58.16 & \bf 16.74 &  41.33 & \bf 56.56 & 84.33 & \bf 73.67 & \bf 64.58 & 4.98 & \bf 12.85 & \bf 18.50 & 44.54 & 39.51 & 35.80 & 9.31 & 22.96 & 33.58\\
  Bagging      & 81.79 & 69.05 & 57.63 & 16.55 & 41.03 & 56.02 & 84.36 & 73.02 & 63.98 & 4.97 & 12.72 & 18.30 & 44.34 & 39.39 & 35.78 & 9.23 & 22.85 & 33.54\\
  Boosted Bagging & 82.69 & \bf 69.68 & 58.14 & 16.73 & \bf 41.43 & \bf 56.56 & \bf 84.55 & 73.58 & 64.53 & \bf 5.00 & 12.82 & 18.47 & \bf 44.64 & \bf 39.65 & \bf 35.96 & \bf 9.32 & \bf 23.03 & \bf 33.72\\ \hline
    Dataset & \multicolumn{6}{c}{\textsf{AmazonCat-13K}} & \multicolumn{6}{c|}{\textsf{Wiki-500K}} & \multicolumn{6}{c}{\textsf{Amazon-3M}} \\ \hline 
    Single Model  & 92.60 & 78.43 & 63.79 & 26.17 & 59.37 & 74.66& 67.05 & 48.02 & \bf 37.52 & 21.84 & 39.85 & \bf 48.08 & 46.42 & 43.60 & 41.52 & 2.89 & 7.17 & 10.59\\\hline
    MC Dropout  & 92.61 & 78.44 & 63.79 & 26.17 & 59.38 & 74.67 & 67.01 & 47.97 & 37.47 & 21.81 & 39.81 & 48.02 & 46.42 & 43.58 & 41.48 & 2.88 & 7.16 & 10.57\\
    Boosting  & 93.01 & 78.81 & \bf 64.08 & 26.30 & 59.68 & \bf 75.02 & \bf 67.85 & \bf 48.06 & 37.33 & \bf 22.17 & \bf 39.98 & 47.96 & 47.05 & \bf 44.29 & \bf 42.16 & \bf 3.03 & \bf 7.48 & \bf 11.01\\
    Bagging & 92.77 & 78.51 & 63.87 & 26.24 & 59.43 & 74.75 & 66.92 & 47.89 & 37.43 & 21.77 & 39.74 & 47.98 & 46.64 & 43.76 & 41.66 & 2.90 & 7.18 & 10.61\\
    Boosted Bagging   & \bf 93.09 & \bf 78.82 & \bf 64.08 & \bf 26.34 & \bf 59.69 & \bf 75.02 & 67.69 & 48.01 & 37.33 & 22.08 & 39.92 & 47.97 & \bf 47.11 & \bf 44.29 & 42.14 & 3.02 & 7.43 & 10.93\\ \hline
\end{tabular}}

\caption{Predictive performance of different methods in percentage (\%) over six experimental datasets. P@$k$ and R@$k$ represent precision and recall metrics with top-$k$ predicted labels.}
    \label{table:acc}
\end{table*}

\subsection{Task 1: Predictive Performance}
\label{section:exp:task1}

Table~\ref{table:acc} shows the predictive performance on six experimental datasets.
Compared to the single model, \emph{MC Dropout} does not improve the predictive accuracy.
This could be because of sparse feature representations in XMC so that dropping model weights is less likely to affect inference.
In contrast, our ensemble approaches outperform using only a single model in most of the metrics since we consider the uncertainty from both data and model perspectives.
For example, \emph{Boosted Bagging} outperforms both Single Model and \emph{MC Dropout} by 1.5\% and 4.8\% in P@1 and R@1 for the {\sf Amazon-3M} dataset.
An interesting observation is: although \emph{Boosted Bagging} performs better for smaller datasets, \emph{Boosting} becomes the best while it comes to larger datasets, such as {\sf Wiki-500K} and {\sf Amazon-3M}.
The reason could be fewer instances, i.e., less knowledge, in training data for each individual model, so those weak models could be incapable of constructing a strong ensemble.

\subsection{Task 2: Misclassification Detection}
\label{section:exp:task2}

The task of misclassification detection aims to evaluate the quality of estimated label-level uncertainty.
Table~\ref{table:labelunc} demonstrates the performance of all methods in average AUROC over six experimental datasets.
All ensemble-based models outperform \emph{Energy} using single model using only entropy as the uncertainty indicator, showing the importance of sampling different model parameters in the manner of Bayesian ensembles~\cite{Gal2016Uncertainty, malinin2020uncertainty, malinin2019uncertainty}.
Our ensemble approaches still outperform all baseline methods over all datasets in misclassification detection.
For instance, PV using \emph{Boosted Bagging} outperforms \emph{Energy} and \emph{MC Dropout} by 2.5\% and 3.4\%.
The observation on performance among our ensemble approaches over different dataset sizes as described in the task of predictive performance still holds here.
We also notice that PV performs better than TU and KU for smaller datasets while TU and KU work better for larger ones.
It can be because the predicted probabilities could be more accurate and less likely to require calibration with more training data.

\begin{table}
\resizebox{\textwidth}{!}{
\centering 
\begin{tabular}{c|c|c|ccc|ccc|ccc}
\hline
    \multirow{2}{*}{Dataset} & \multirow{2}{*}{Energy} & MC & \multicolumn{3}{c|}{Boosting}  & \multicolumn{3}{c|}{Bagging} &  \multicolumn{3}{c}{Boosted Bagging}\\ \cline{4-12}
 &  & Dropout  & PV & TU & KU & PV & TU & KU & PV & TU & KU\\ \hline
 \textsf{Eurlex-4K} & 83.54 & 93.54 & 93.37 & 93.51 & 94.34 & 94.19 & 93.47 & 93.73 & \bf 94.39 & 93.53 & 94.18\\
 \textsf{Wiki10-31K} & 87.28 & 86.48 & 86.44 & 87.36 & 87.28 & 86.30 & 87.14 & 87.09 & \bf 89.46 & 87.18 & 87.14\\
 \textsf{Amazon-670K} & 80.73 & 95.94 & 95.02 & 95.65 & 95.55 & \bf 96.12 & 95.67 & 95.67 & 95.82 & 95.64 & 95.49\\
 \textsf{AmazonCat-13K} & 85.15 & 95.50 & \bf 95.87 & 95.58 & 95.47 & 95.68 & 95.52 & 95.53 & 95.35 & 95.73 & 95.70\\
 \textsf{Wiki-500K} & 80.10 & 93.59 & 92.15 & 93.42 & \bf 94.07 & 93.95 & 93.52 & 93.76 & 93.33 & 93.43 & 93.95\\
 \textsf{Amazon-3M} & 77.04 & 77.00 & 75.46 & \bf 77.42 & 75.18 & 76.03 & 76.95 & 76.36 & 75.87 & 77.26 & 76.03\\\hline
\end{tabular}}

\caption{The misclassification detection performance of different methods in average AUROC (\%) for evaluating the quality of label-level uncertainty. PV, TU, and KU represent \emph{Probability Variance}, \emph{Total Uncertainty}, and \emph{Knowledge Uncertainty}.}

    \label{table:labelunc}
\end{table}

\subsection{Task 3: Out-of-Distribution (OOD) Detection}
\label{section:exp:task3}

In the task of OOD detection, we evaluate the quality of estimated instance-level uncertainty.
Table~\ref{table:instunc} shows the performance in AUROC for all methods with six datasets.
Similar to the experimental results in misclassification detection, ensemble-based methods outperform \emph{JointEnergy} using single models in most cases.
Another similar observation is: with the smaller training dataset like {\sf Wiki10-31K} and smaller model sizes, \emph{MC Dropout} and \emph{Bagging} perform better.
On the other hand, \emph{Boosting} and \emph{Boosted Bagging} are the best methods when it comes to using {\sf Amazon-670K} as the training dataset.
Moreover, general performance on distinguishing {\sf AmazonCat-13K} and {\sf Amazon-3M} from {\sf Amazon-670K} is lower because of similar dataset distributions.
However, our proposed methods can still outperform baseline methods.
We further found that variance-based quantification methods (i.e., \emph{MC Dropout} and PV) generally perform better than entropy-based approaches.

\begin{table*}
\centering
\resizebox{\textwidth}{!}{
\begin{tabular}{c|c|c|c|ccc|ccc|ccc}
\hline
\multirow{2}{*}{Training Dataset} & \multirow{2}{*}{OOD Dataset} & Joint & MC & \multicolumn{3}{c|}{Boosting} & \multicolumn{3}{c|}{Bagging} & \multicolumn{3}{c}{Boosted Bagging} \\ \cline{5-13}
 &  & Energy  & Dropout  & PV & TU & KU & PV & TU & KU & PV & TU & KU\\ \hline
\multirow{5}{*}{\textsf{Wiki10-31K}} & \textsf{Eurlex-4K} & 96.89 & 97.11 & 97.17 & 96.71 & 95.76 & \bf 97.21 & 96.97 & 96.76 & 97.08 & 96.81 & 96.58\\
& \textsf{Amazon-670K} & 95.55 & \bf 96.66 & 96.53 & 95.52 & 95.13 & 96.58 & 95.68 & 95.39 & 96.56 & 95.70 & 95.35\\
& \textsf{AmazonCat-13K} & 94.60 & \bf 95.91 & 95.76 & 94.63 & 93.99 & 95.82 & 94.75 & 94.45 & 95.82 & 94.81 & 94.37\\
& \textsf{Wiki-500K} & 91.01 & 90.43 & 90.70 & 90.71 & 89.10 & 90.91 & 91.12 & \bf 91.19 & 90.92 & 91.09 & 91.13\\
& \textsf{Amazon-3M} & 94.89 & \bf 96.28 & 96.17 & 94.87 & 94.31 & 96.14 & 95.06 & 94.63 & 96.15 & 95.10 & 94.48\\ \hline
\multirow{5}{*}{\textsf{Amazon-670K}} & \textsf{Eurlex-4K} & 98.68 & 98.67 & 98.79 & \bf 98.85 & 98.65 & 98.75 & 98.62 & 98.71 & 98.83 & 98.76 & 98.69\\
& \textsf{Wiki10-31K} & 71.01 & 72.85 & 72.83 & 71.27 & 70.90 & 73.22 & 71.28 & 71.21 & \bf 73.31 & 71.38 & 71.26\\
& \textsf{AmazonCat-13K} & 48.38 & 50.62 & \bf 51.11 & 48.41 & 49.50 & 50.63 & 48.39 & 48.34 & 50.63 & 48.17 & 48.84\\
& \textsf{Wiki-500K} & 88.29 & 88.78 & 88.55 & 88.64 & 88.04 & 89.27 & 88.36 & 88.50 & \bf 89.58 & 88.88 & 88.43\\
& \textsf{Amazon-3M} & 64.27 & 66.48 & \bf 67.15 & 65.08 & 64.58 & 66.67 & 64.31 & 64.43 & 66.59 & 64.49 & 64.55\\ \hline 

\end{tabular}}

\caption{The out-of-distribution (OOD) detection performance of different methods in AUROC (\%) for evaluating the quality of instance-level uncertainty. PV, TU, and KU represent \emph{Probability Variance}, \emph{Total Uncertainty}, and \emph{Knowledge Uncertainty}.}
    \label{table:instunc}
\end{table*}

\subsection{Approximation Efficiency}
\label{section:exp:efficiency}

Table~\ref{table:efficiency} states the execution time with the performance of Tasks 2 and 3 for \emph{Boosted Bagging} based on the na\"ive approach and our beam search approximation for uncertainty quantification.
As a result, beam search approximation is significantly faster than the na\"ive approach.
For example, in misclassification detection, our approximation obtains 110.39x and 30.32x speedups with only 1.7\% and 3.0\% drop in AUROC using PV on \wikis and \eurlex datasets .
In OOD detection, after training XMC models with the \wikis dataset, beam search is 61.66x and 39.78x faster than the na\"ive method with only 0.1\% and 0.5\% performance loss in AUROC using PV to identify \eurlex and \amzsmall as OOD datasets.

\begin{table}[!t]
\centering
\begin{tabular}{c|c|c|ccc}
    \hline
    & \multirow{2}{*}{Approach} & Time & \multicolumn{3}{c}{Boosted Bagging} \\ \cline{4-6}
    & & (Minutes) & PV & TU & KU \\ \hline 
    Dataset & \multicolumn{5}{c}{Misclassification Detection} \\ \hline
    \multirow{2}{*}{\eurlex}
    & Na\"ive &  49.12 & 97.35 & 97.17 & 97.23\\ 
    & Beam Search &\bf 1.62 (30.32x) & 94.39 & 93.53 & 94.18 \\ \hline
    \multirow{2}{*}{\wikis}
    & Na\"ive & 671.18 & 91.06 & 90.77 & 91.04 \\ 
    & Beam Search &\bf 6.08 (110.39x) & 89.46 & 87.18 & 87.14 \\ \hline
    OOD Dataset & \multicolumn{5}{c}{OOD Detection (Trained on \wikis)} \\ \hline 

    \multirow{2}{*}{\eurlex}
    & Na\"ive &  1135.82 & 97.13 &  97.10 &  96.91 \\
    & Beam Search & \bf 18.42 (61.66x) & 97.08 &  96.81 &  96.58 \\ \hline
    
    \multirow{2}{*}{\amzsmall}
    & Na\"ive & 2501.06 & 97.02 &  96.16 &  96.28 \\ 
    & Beam Search & \bf 62.87 (39.78x) & 96.56 &  95.70 &  95.35 \\ \hline 

\end{tabular}

\caption{The execution time with the performance of misclasification detection and OOD detection for the \emph{Boosted Bagging} method based on the na\"ive approach and beam search approximation for uncertainty quantification. PV, TU, and KU represent \emph{Probability Variance}, \emph{Total Uncertainty}, and \emph{Knowledge Uncertainty}.}
    \label{table:efficiency}

\end{table}

\subsection{Performance on Deep XMC Models}
\label{section:exp:deep}

To further demonstrate the potential of our proposed approach with deep learning, we adopt \xrtransformer, which is the state-of-the-art XMC model based text data and transformers~\cite{devlin2018bert,vaswani2017attention}, as the base XMC model.
Table~\ref{table:deepacc} shows the predictive performance with \xrtransformer as the base XMC model on three datasets.
Our ensemble-based method using \emph{Boosted Bagging} outperforms the single \xrtransformer model across all metrics.
For evaluating the quality of estimated uncertainty, Table~\ref{table:deeplabelunc} provides the performance on misclassification detection.
All uncertainty metrics based on \emph{Boosted Bagging} are better than \emph{Energy} using a single model.
Besides, \xrtransformer with uncertainty quantification using our approach can actually further yield better state-of-the-art results.
Table~\ref{table:alldeepacc} shows the predictive performance of various XMC models and our approach \emph{Boosted Bagging} with \xrtransformer as the base XMC model, where comparative methods do not consider uncertainty quantification.
As a result, after applying our ensemble-based uncertainty quantification approach to \xrtransformer, we can obtain state-of-the-art performance in most of the metrics.
This further demonstrate the benefits of modeling uncertainty in the XMC task.

\begin{table}[!t]
\centering
\begin{tabular}{c|cccccc}
\cline{2-7}
    & P@1 & P@3 & P@5 & R@1 & R@3 & R@5 \\ \hline 
 Dataset & \multicolumn{6}{c}{\eurlex} \\ \hline
 Single Model & 87.17 & 74.51 &  61.51 &  17.71 &  44.45 &  59.86\\
 Boosted Bagging & \bf 88.10 & \bf  75.71 & \bf  62.21 & \bf  17.92 & \bf  45.10 & \bf  60.53 \\ \hline
 Dataset & \multicolumn{6}{c}{\wikis} \\ \hline
 Single Model & 87.89 & 78.69 &  69.08 &  5.24 &  13.84 &  19.89\\
 Boosted Bagging & \bf 88.54 & \bf  79.68 & \bf  70.31 & \bf  5.29  & \bf 14.01 & \bf  20.24 \\ \hline
 Dataset & \multicolumn{6}{c}{\amzsmall} \\ \hline
    Single Model & 49.00 & 43.68 & 39.81 & 10.30 & 25.49 & 37.49 \\ 
 Boosted Bagging & \bf 49.95 & \bf  44.58 & \bf  40.67 & \bf  10.49  & \bf 25.99 & \bf  67.26 \\ \hline
\end{tabular}

\caption{Predictive performance with \xrtransformer as the base XMC model in percentage (\%) on three datasets.}
 \label{table:deepacc}
\end{table}

\begin{table}[!t]
\centering
\begin{tabular}{c|c|ccc}

\hline
\multirow{2}{*}{Dataset} & \multirow{2}{*}{Energy} & \multicolumn{3}{c}{Boosted Bagging}\\ \cline{3-5}
& & PV & TU & KU \\ \hline 
    \eurlex & 90.10 & 91.74 &\bf 92.75 & 92.03 \\
    \wikis & 88.19 & 88.29 &\bf 89.28 & 88.69 \\ 
    \amzsmall & 91.30 & 92.46 & \bf 95.04 & 93.10 \\ \hline
\end{tabular}

\caption{The misclassification detection performance with \xrtransformer as the base XMC model on three datasets. PV, TU, and KU represent \emph{Probability Variance}, \emph{Total Uncertainty}, and \emph{Knowledge Uncertainty}.}
\label{table:deeplabelunc}
\end{table}

\begin{table*}[!t]
\centering
\resizebox{\linewidth}{!}{
\begin{tabular}{c|ccc|ccc|ccc}
\cline{2-10}
    & P@1 & P@3 & P@5 & P@1 & P@3 & P@5 & P@1 & P@3 & P@5 \\ \hline 
 Method & \multicolumn{3}{c|}{\eurlex} & \multicolumn{3}{c|}{\wikis}  & \multicolumn{3}{c}{\amzsmall}\\ \hline
    AnnexML~\cite{tagami2017annexml} &  79.66 &  64.94 &  53.52 &  86.46 &  74.28 &  64.20 & 42.09 &  36.61 &  32.75 \\
    DiSMEC~\cite{babbar2017dismec} &  83.21 &  70.39 &  58.73 &  84.13 &  74.72 & 65.94 & 44.78 &  39.72 &  36.17\\
    PfastreXML~\cite{jain2016extreme} &  73.14 &  60.16 &  50.54 &  83.57 &  68.61 &  59.10 & 36.84 &  34.23 &  32.09 \\
    Parabel~\cite{prabhu2018parabel} &  82.12 &  68.91 &  57.89 &  84.19 &  72.46 &  63.37 & 44.91 &  39.77 &  35.98 \\
    eXtremeText~\cite{wydmuch2018no} &  79.17 &  66.80 &  56.09 &  83.66 &  73.28 &  64.51 & 42.54 &  37.93 &  34.63 \\
    Bonsai~\cite{khandagale2020bonsai} &  82.30 &  69.55 &  58.35 &  84.52 &  73.76 &  64.69 & 45.58 &  40.39 &  36.60 \\
    XML-CNN~\cite{liu2017deep} &  75.32 &  60.14 &  49.21 &  81.41 &  66.23 &  56.11 & 33.41 &  30.00 &  27.42 \\
    AttentionXML~\cite{you2019attentionxml} &  85.49 &  73.08 &  61.10 &  87.05 &  77.78 &  68.78 & 45.66 &  40.67 &  36.94 \\
    X-Transformer~\cite{chang2020xmctransformer} &  85.82 &  73.23 & 61.18 &  87.79 &  78.43 &  69.74  &46.45 & 40.82 & 36.71\\ 
    LightXML~\cite{jiang2021lightxml} &  85.60 & 74.10 & 62.00 & 87.84 & 77.26 &67.99 & 47.30 &  42.20 &  38.50 \\
    \xrtransformer~\cite{zhang2021fast} & 87.17 & 74.51 &  61.51 &  87.89 & 78.69 &  69.08 & 49.00 & 43.68 & 39.81\\ \hline
    \xrtransformer with \emph{Boosted Bagging} & \bf 88.10 & \bf  75.71 & \bf  62.21 & \bf 88.54 & \bf  79.68 & \bf  70.31 & \bf 49.95 & \bf 44.58 & \bf 40.67\\ \hline
\end{tabular}
}

\caption{Predictive performance of various XMC models and our approach using \xrtransformer as the base XMC model on \eurlex and \wikis. P@$k$ represents precision with top-$k$ predicted labels.}
 \label{table:alldeepacc}
\end{table*}

\section{Related Work}
\label{section:relatedwork}

\mysection{Ensemble-based Uncertainty Quantification}
Uncertainty quantification aims at characterizing (and reducing) uncertainties in computational applications~\cite{walker2003defining}.
Instead of using a single model capturing only \emph{data uncertainty} (or aleatoric uncertainty) ~\cite{Gal2016Uncertainty, malinin2019uncertainty, duan2020ngboost}, ensemble-based approaches generate ensembles of models so that the overall estimated uncertainty can be decomposed into \emph{data uncertainty} and \emph{knowledge uncertainty} (or epistemic uncertainty)~\cite{depeweg2018decomposition}.
For example, \citet{gal2016dropout} collect model ensembles by dropping model weights of neural networks during training;
\citet{malinin2020uncertainty} generate ensembles of gradient boosting decision trees with intermediate individual models;
\citet{malinin2020uncertainty_struct} apply the entropy chain rule and sampling to collect an ensemble for autoregressive structured prediction;
Deep Ensemble~\cite{lakshminarayanan2017simple} derives the model ensemble by varying initial weights of neural models.
These ensemble-based uncertainty quantification methods have been applied to real-world applications in various domains, such as medicine diagnosis~\cite{roy2019bayesian, dahal2020uncertainty} and computer vision~\cite{zhang2020mix, liu2019variable}.
In addition, ensemble models are also usually capable of improving predictive performance.
However, although ensemble-based approaches have obtained many successes in different fields, none of the previous studies focuses on the XMC task due to the challenges introduced in Section~\ref{section:intro}.

\mysection{Partition-based Extreme Multi-label Classification}
Computational challenge is always the hurdle of the XMC problem.
Even with sparse linear XMC models~\cite{babbar2019data} comparatively more lightweight, na\"ive one-versus-rest (OVR) methods~\cite{yen2016pd} with a linear inference time could be still too slow to be applied in real-world applications.
Partition-based methods are one of the most popular approaches to addressing the efficiency and scalability of XMC models~\cite{chang2021extreme}.
By introducing different partitioning techniques, the enormous label spaces can be reduced into hierarchical label trees so that the exhausting process on examining all labels can be replaced by efficient semantic indexing as mentioned in Section~\ref{section:prelim:xmc}.
Parabel~\cite{prabhu2018parabel} first establishes balanced 2-means label trees with instance-induced label features.
Accordingly, various successors propose diverse partitioning and indexing methods for improvements, such as eXtremeText~\cite{wydmuch2018no}, Bonsai~\cite{khandagale2020bonsai}, and PECOS~\cite{yu2020pecos}.
Moreover, with longer training and inference time,  partition-based methods can obtain state-of-the-art accuracy by utilizing deep neural encoders, such as AttentionXML~\cite{you2019attentionxml}, LightXML~\cite{jiang2021lightxml}, and \xrtransformer~\cite{zhang2021fast}, for time-insensitive applications.
In particular, \xrlinear in PECOS and \xrtransformer are the state-of-the-art sparse linear and deep XMC models, and treated as the base XMC models in our experiments.
For simplicity, partition-based methods are called \emph{tree-based methods} in this work.

\section{Conclusions}
\label{section:conclusions}

In this paper, we are the pioneer of studying uncertainty quantification for eXtreme Multi-label Classification (XMC).
We first propose to generate ensembles with the techniques of bootstrapping and boosting for estimating uncertainty from different perspectives..
Besides, we also come up with two different uncertainty levels, including \emph{label-level} and \emph{instance-level} uncertainty.
To overcome the computational issues over the enormous label spaces, we further suggest to approximating uncertainty with beam search under long-tail probability distribution in XMC problems.
In experiments, we demonstrate our proposed approaches not only obtain superior predictive performance, but also result in high-quality uncertainty estimation in both label-level and instance-level uncertainty, compared to baseline methods.
Moreover, our framework can deliver better state-of-the-art XMC results after considering uncertainty quantification and using deep XMC models as base models.

\clearpage

\appendix

\section*{Appendix}
\section{Proof of Theorem~\ref{theorem:beamsearch}}
\label{appendix:theorem:beamsearch}


\begin{proof}
    Without loss of generality, we assume a pair of $(o_i, \ell_i)$ resulting a positive regret, where $o_i$ is out of beam search results; $P(y^t_{o_i}\mid \bx; \btheta) >  P(y^t_{\ell_i}\mid \bx; \btheta)$. Based on Equation~\eqref{eq:treerec}, we have:
    \begin{equation*}
        P(y^t_{o_i}\mid \bx; \btheta) = P(y^t_{o_i}\mid \bx, c^t_{o_i}; \btheta) \times P(c^t_{o_i}\mid \bx; \btheta),
    \end{equation*}
    where $c^t_{o_i}$ is the corresponding cluster in the previous layer. Since $o_i$ is out of beam search results, $o_i$ is ranked after the $k_{t-1}$-th position among $P(y^{t-1}\mid \bx;\btheta)$.
    As mentioned in Section~\ref{section:prelim:xmc}, $P(c^t\mid \bx; \btheta)$ represents $P(y^{t-1}\mid \bx; \btheta)$ in the recursive manner of tree-based XMC models, so $P(c^t\mid \bx; \btheta)$ is also under a long-tail distribution.
    Hence, if the beam size $k_{t-1}$ is large enough,  we would have:
    \begin{equation*}
        P(c^t_{o_i}\mid \bx; \btheta) \leq \delta^\prime, \text{ where } \delta^\prime\approx 0 \text{ is a small positive constant}.
    \end{equation*}
    Moreover, the probability $P(y^t_{o_i}\mid \bx; \btheta)$ is also bounded as:
    \begin{equation*}
        P(y^t_{o_i}\mid \bx; \btheta) \leq P(y^t_{o_i}\mid \bx, c^t_{o_i}; \btheta) \times \delta^\prime \leq \delta^\prime.
    \end{equation*}
    Based on the assumption , we then have:
    \begin{equation} \label{eq:beamsearch:loc}
        P(y^t_{o_i}\mid \bx; \btheta) -  P(y^t_{\ell_i}\mid \bx; \btheta) \leq \delta^\prime.
    \end{equation}
    By extending Equation~\eqref{eq:beamsearch:loc} to all positions, we would have:
    \begin{equation*}
        \frac{1}{k^t} \sum_{i=1}^{k^t} (P(y^t_{o_i}\mid \bx; \btheta) -  P(y^t_{\ell_i}\mid \bx; \btheta)) \leq \frac{1}{k^t} \sum_{i=1}^{k^t} \delta^\prime = \delta^\prime.
    \end{equation*}

\end{proof}

\vspace{-24pt}
\section{Proof of Theorem~\ref{theorem:approx}}
\label{appendix:theorem:approx}

For simplicity, in our proofs, $\hat{P}(y_i\mid \bx, \btheta^{(m)})$ represents the $i$-th top probability approximated by beam search in Algorithm~\ref{alg:approxbeam} as:
    \begin{equation*}
    \hat{P}(y_i\mid \bx, \btheta^{(m)}) = \left\lbrace
        \begin{array}{cl}
            P(y_\ell\mid \bx; \btheta^{(m)}) & , \text{ if } i \leq k, \\
            \delta &, \text{ else}
        \end{array}\right., \text{ where $\delta\approx 0$.}
    \end{equation*}
Here we start from proving the following Lemma~\ref{lemma:longtaildiff}.

\begin{lemma} \label{lemma:longtaildiff}
    For each set of model parameters $\btheta^{(m)}$, suppose $P(y_i\mid \bx; \btheta^{(m)})$ denotes the $i$-th greatest probability in $\{ P(y_\ell \mid \bx; \btheta^{(m)})\mid \ell \in \cL\}$. If probabilities $P(y_\ell \mid \bx; \btheta^{(m)})$ is under a long-tail distribution as shown in Figure~\ref{fig:longtail}, with a large enough number $k$, for any label $\ell$ we have:
    \begin{equation*} 
        \left|(-\ln P(y_\ell\mid \bx; \btheta^{(m)})) - (-\ln \hat{P}(y_\ell\mid \bx; \btheta^{(m)}))\right| \leq \delta^\prime, 
    \end{equation*},
    where $\delta^\prime \approx 0$ is a small positive constant;  \end{lemma}
\begin{proof}
    Since  $P(y_\ell \mid \bx; \btheta^{(m)})$ is under a long-tail distribution as shown in Figure~\ref{fig:longtail}, with a large enough $k$, we have:
    \begin{equation*}
        P(y_i \mid \bx; \btheta^{(m)}) \approx 0 \approx \delta, \forall i> k.
    \end{equation*}
    For the top cases, where $i\leq k$, it is trivial that $|(-\ln P(y_\ell\mid \bx; \btheta^{(m)})) - (-\ln \hat{P}(y_\ell\mid \bx; \btheta^{(m)}))| = 0 \leq \delta^\prime$ for any $\delta^\prime > 0$.
    For the other cases, i.e., $i > k$, we have:
    \begin{align*}
        |( & -\ln p(y_\ell\mid \bx; \btheta^{(m)})) - (-\ln \hat{p}(y_\ell\mid \bx; \btheta^{(m)}))| \\
        &= \left|\ln \hat{P}(y_\ell\mid \bx; \btheta^{(m)}) - \ln P(y_\ell\mid \bx; \btheta^{(m)}) \right|  \\
        &= \left|\ln \frac{\hat{P}(y_\ell\mid \bx; \btheta^{(m)})}{P(y_\ell\mid \bx; \btheta^{(m)})} \right| = \left|\ln \frac{\delta}{P(y_\ell\mid \bx; \btheta^{(m)})} \right| \approx \left|\ln \frac{\delta}{\delta} \right| = \left| \ln 1 \right| = 0.
    \end{align*}
    Therefore, it must exist a small positive constant $\delta^\prime$ so that $|(-\ln P(y_\ell\mid \bx; \btheta^{(m)})) - (-\ln \hat{P}(y_\ell\mid \bx; \btheta^{(m)}))| \leq \delta^\prime$.
\end{proof}

Based on Lemma~\ref{lemma:longtaildiff}, we can prove Theorem~\ref{theorem:approx} accordingly.

\begin{proof}
    \begin{align*}
       |& U(\bx, \ell) - \hat{U}(\bx, \ell)| = \left|\cH[P(y_\ell\mid \bx, \cD)] - \cH[\hat{P}(y_\ell \mid \bx, \cD)]\right| \\
        &= \left|\bbE_{p(y_\ell\mid \bx, \cD)}[-\ln P(y_\ell\mid \bx, \cD) - \bbE_{p(y_\ell\mid \bx, \cD)}[-\ln \hat{P}(y_\ell\mid \bx, \cD) ] \right|\\
        &= \left|\bbE_{p(y_\ell\mid \bx, \cD)}[(-\ln P(y_\ell\mid \bx, \cD)) - (-\ln \hat{P}(y_\ell\mid \bx, \cD)) ]\right|\\
        &\approx \left| \frac{1}{M} \sum_{m=1}^M \left( (-\ln P(y_\ell\mid \bx; \btheta^{(m)})) - (-\ln \hat{P}(y_\ell\mid \bx; \btheta^{(m)})) \right) \right|\\
        &\text{(By ensemble-based approximation)}\\
        &\leq \frac{1}{M} \sum_{m=1}^M \left| \left( (-\ln P(y_\ell\mid \bx; \btheta^{(m)})) - (-\ln \hat{P}(y_\ell\mid \bx; \btheta^{(m)})) \right) \right|\\
        &\text{(By the Minkowski inequality)} \leq  \frac{1}{M} \sum_{m=1}^M \delta^\prime  = \delta^\prime \text{~(By Lemma~\ref{lemma:longtaildiff})}
    \end{align*}
\end{proof}

\bibliographystyle{abbrvnat}
\bibliography{main}

\end{document}